# Gabor Surface Feature for Face Recognition

Ke Yan, Youbin Chen
Graduate School at Shenzhen
Tsinghua University
Shenzhen, China
xjed09@gmail.com, chenyb@sz.tsinghua.edu.cn

David Zhang
Department of Computing
Hong Kong Polytechnic University
Hong Kong, China
csdzhang@comp.polyu.edu.hk

*Abstract*—Gabor filters can extract multi-orientation and multi-scale features from face images. Researchers have designed different ways to use the magnitude of the filtered results for face recognition: Gabor Fisher classifier exploited only the magnitude information of Gabor magnitude pictures (GMPs); Local Gabor Binary Pattern uses only the gradient information. In this paper, we regard GMPs as smooth surfaces. By completely describing the shape of GMPs, we get a face representation method called Gabor Surface Feature (GSF). First, we compute the magnitude, $1^{st}$ and $2^{nd}$ derivatives of GMPs, then binarize them and transform them into decimal values. Finally we construct joint histograms and use subspace methods for classification. Experiments on FERET, ORL and FRGC 1.0.4 database show the effectiveness of GSF.

## I. INTRODUCTION

Face recognition has attracted significant attention due to its wide applications. As a typical pattern recognition problem, face recognition includes two main steps: feature extraction and classification. Since the appearance of the same face could look dramatically different when captured in different environments, it is very important to extract features robust to variations such as illumination, time, pose and expression.

Many features have been proposed for face recognition. Among them, Gabor wavelet shows very promising performance [1]. Due to the high dimensionality of Gabor features, subspace methods, such as Fisherface, have been applied with it, forming the Gabor Fisher Classifier (GFC) [2].

Local Binary Pattern (LBP) [3] was firstly designed for texture classification. Ahonen et al. [4] successfully applied it to represent faces. Their results show that facial images can be seen as a composition of micro-patterns, such as flat areas, spots, lines and edges. Huang et al. [5] ran LBP operator on gradient magnitude images so that the second derivative information is extracted. Guo et al. [6] used the magnitude of the central pixel together with the sign and magnitude of the gradient to build a 3D joint histogram for texture classification, obtaining better results than the origin LBP.

The link between Gabor feature and LBP are not established until Zhang et al.'s study [7]. They found that one could firstly use multi-scale and multi-orientation Gabor filters to decompose a face image into Gabor Magnitude Pictures (GMPs), followed by the LBP operator. This combination, known as Local Gabor Binary Pattern (LGBP), is more robust and with more discriminating power than previous features like LBP, but it suffers the disadvantage of high dimensionality. To solve this problem as well as improve the recognition performance, their team further proposed a statistical extension for LGBP similarity computation by introducing Fisher Discriminant Analysis (FDA) to the LGBP features [8]. This improvement got impressive recognition score on the standard FERET database.

The success of LGBP brings us one question: how does LGBP work? As we all know, LBP is originally a texture extractor that detects micro-patterns. However, does it work the same way when it is applied on GMPs? By analyzing this, we may be able to come up with a novel operator to replace LBP, and expect it to have more explicit "physical meanings", meanwhile show better performance.

In this paper, we propose to treat a GMP as a smooth surface. The distinctive information is hidden in its shape, which is not only their magnitude values, but also $1^{st}$ and $2^{nd}$ derivatives. LBP actually contains the first derivative information of GMPs. Properly introducing magnitude and second derivative information will intuitively bring better performance. Our method calculates all these values from GMPs, binarizes them, and then forms a joint multi-dimensional histogram. We call this a Gabor Surface Feature (GSF). After that, we use Ensemble of Piecewise FDA (EPFDA) classifiers (proposed in [8]) to do the final classification. Experiments on FERET database, ORL and Face Recognition Grand Challenge version 1 experiment 1.0.4 ('FRGC-104') show the effectiveness of GSF. On FERET we have got an impressive recognition rate.

This paper is organized as follows: Section II briefly reviews LGBP and EPFDA. Section III presents the GSF scheme. Section IV reports experimental results and Section V concludes the paper.

## II. BRIEF REVIEW OF LGBP AND EPFDA

It is actually very easy to describe the LGBP algorithm. First, we use multi-orientation and multi-scale Gabor filters (usually it is 8 orientations and 5 scales) to convolve a face image and keep the magnitudes. Second, we apply $LBP_{8,1}$ operator [3] on these 40 GMPs, turning them into 40 LGBP maps. Third, in order to reduce the length of the LGBP histogram, we need to quantize the LBP codes into $L$ levels; a typical value of $L$ is 8. After that, we exploit local feature histogram to summarize the region property of the LGBP patterns. Each LGBP map is divided into $M*N$ non-overlapped regions, from which histograms are computed. Then $M*N*40$ histograms are concatenated to a histogram sequence (LGBPHS), as the final LGBP face representation.

This work is partially supported by the Natural Science Foundation of China (NSFC) (No. 61101150).

In [7] histogram intersection is adopted to compare two LGBPHSs. So the training stage is unnecessary. The generalizability problem is thus naturally avoided. However, let's set $M = 11$ and $N = 20$, then the feature vector will be as long as 70,400. It will take a large space to store these vectors and a long time for matching. So EPFDA is constructed in [8] to reduce the dimensionality of the vector. In [8], each LGBP map is divided into $M*N$ non-overlapped regions. Then each region is further partitioned into $S$ sub-regions. Compute a histogram for every sub-region, concatenate them within every region, and finally we can build an FDA classifier for each region. Now we have $M*N$ FDA subspaces. For each of them, the input dimension is $40*S*L$, the reduced dimension is $R$.

When matching two faces, we first compute the origin LGBPHS $V$,

$$V = (X_1, X_2, \cdots, X_{M*N}). \quad (1)$$

Then we use the $M*N$ FDA matrices to transform $V$ into a low-dimensional representation $F$:

$$F = (F_1, F_2, \cdots, F_{M*N}), \quad (2)$$

$$F_j = (W_j^{FLD})^T X_j. \quad (3)$$

The total matching score for two faces $V$ and $V'$ can be written as:

$$\text{Score}(V, V') = \sum_{j=1}^{M*N} s(F_j, F'_j), \quad (4)$$

where s is the cosine metric. Experiments in [8] prove that:

- FDA classifier can not only reduce the dimensionality of the histogram sequence vector, but also make it more discriminating.

- Building piecewise FDA classifiers and then fusing them with a sum rule is a good idea to preserve the information of every region of the face.

Due to the reasons mentioned above, we will adopt the EPFDA strategy after we extract GSF features on faces.

## III. FACE REPRESENTATION BASED ON GABOR SURFACE FEATURE

### A. Motivation and Basic Ideas of GSF

Before introducing GSF, it's better to get some insight into LGBP. Because of the Gaussian envelope of the Gabor kernel, the GMPs look like smooth surfaces, as shown in Fig. 1.There is obviously no "micro-patterns" on GMPs. So why would LBP work well on GMPs? Remind that the nature of LBP is computing the gradient of a pixel in all orientations. We could safely guess that applying LBP can extract the gradient information on GMPs. For example, the 3 LBP operators in Fig. 2 can extract a peak, a slope in +x orientation, and a valley, respectively.

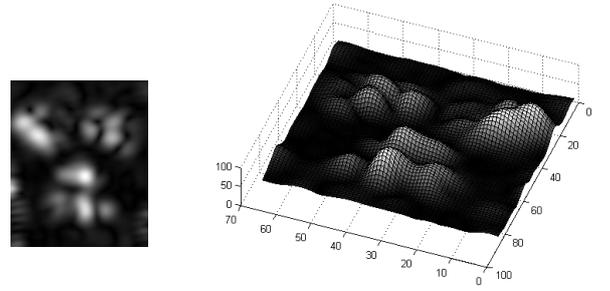

Figure 1. 2D and 3D representation of a GMP

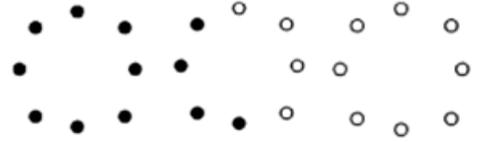

Figure 2. Examples of LBP operators. A solid circle means the pixel is darker than the central pixel. A hollow one means the opposite

From this point of view, we can aim at designing a feature that better summarizes the properties on GMPs. Note that a GMP has not only gradient, but also magnitude and second derivative. We would like to include them into GSF. "Binarization and joint histogram" is the basic strategy for many LBP-based operators; the effectiveness of this strategy has been proved massively [6]. Therefore, we would like to use it, too.

What is more, in [7] and [8] there is a quantization step after extracting LGBP features. The quantization seems a little arbitrary, because two LBP codes can be close to each other in decimal format, but not similar in fact, for example, 00000111 and 00001000. Others may be similar in fact but quantized to different bins. We would like our new feature to get rid of this problem.

### B. Gabor Surface Feature

Now we formally introduce the Gabor Surface Feature. Using simple symmetric gradient operators $[-1,0,1]$ and $[-1,0,1]^T$, we can filter a Gabor magnitude picture $G$, and get its gradient picture, $Gx$ and $Gy$. Then we can binarize $G$, $Gx$ and $Gy$, getting $B$, $Bx$ and $By$. The threshold is the median value of $G$, $Gx$ and $Gy$, so as to make the histogram distributed uniformly. The next step is similar to LBP; we combine the binarized pictures to a decimal code map $F$ by:

$$F = 2^2 B + 2^1 B_x + 2^0 B_y. \quad (5)$$

$F$ is an example of our proposed GSF map. It includes both magnitude and gradient information of each pixel. It has only 8 values so no quantization is needed. In order to add second derivative information further, we filter $Gx$ and $Gy$ with gradient operators and get $Gxx$, $Gyy$, and $Gxy$. If we combine them altogether using the same method as in (5), the number of bins will be too large. We tried several ways to combine them and finally found 2 ways with the highest recognition rate:

$$F = 2^3 B + 2^2 B_x + 2^1 B_y + 2^0 B_2, \quad (6)$$

$$F = 2^3 B_x + 2^2 B_y + 2^1 B_{xx} + 2^0 B_{yy}. \quad (7)$$

In (6), $B_2$ is the binarization of $Gxx+Gyy$. It's an indicator of the convexity of the surface. These two methods have both 16 bins. Eq. (6) contains magnitude, 1st and 2nd derivative information. However, it can't defeat (7) when the illumination condition is bad. The reason is probably that the magnitude of a GMP is not robust to illumination variations, compared with the derivatives. Therefore, when no light preprocessing is done before extracting GSF, we will use (7); otherwise, we can expect (6) to get a better recognition rate. The preprocessing algorithm will be briefly introduced in section C. After generating the GSF map, we'll use the histogram and EPFDA strategy described in section II. The EPFDA will be a weighted version, as shown in section D.

The meaning of each GSF value is clear and can be easily found from Fig. 3. We take the 1D situation as an example, here $F_1 = 2^2 B + 2^1 Bx + 2^0 Bxx$. Each value indicates a certain shape of the 1D curve.

Fig. 4 shows a GSF map example according to (6). The picture on the right is the LGBP map of the same GMP. Obviously, the GSF map is more meaningful. In section IV, we will see the performance of GSF is also competitive.

### C. Illumination preprocessing

A major difficulty that face recognition systems will encounter is the varied illumination conditions. A good preprocessing algorithm may reduce the intra-class difference and improve the performance. Here we choose the processing sequence in [9]. The sequence is made of 4 stages: gamma correction, difference of Gaussian filtering, masking and contrast equalization. The masking stage is not used in this paper. We find this algorithm can perfectly remove shadows and preserve details on faces, as shown in Fig. 5.

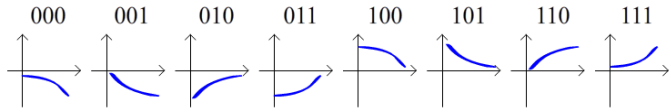

Figure 3. Correspondence between GSF value and curve shape

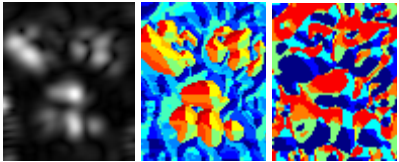

Figure 4. An example of GMP(left) and its GSF(middle), LGBP(right) map

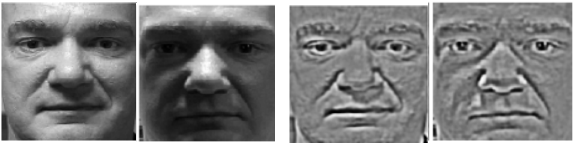

Figure 5. Before preprocessing (left) and after preprocessing (right) [9]

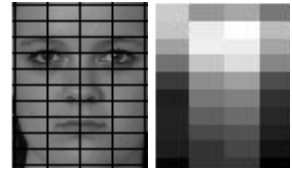

Figure 6. The weights of the different regions of faces

### D. Weighted EPFDA

Weighting is a common strategy in pattern recognition to further improve the performance of an existing algorithm. Considering the fact that each region of the face has different discriminating power, we would like to assign weights on them when calculating the final matching score. Eq. (4) is then modified by:

$$\text{Score}(V,V') = \sum_{j=1}^{M*N} w_j \text{s}(F_j, F_j'), \quad (8)$$

where $w_j$ is the weight. Similar to the method in [4], a training set was classified using only one of the $M*N$ regions at a time, the recognition rate was directly used as the weight of that region. The obtained weights are illustrated in Fig. 6.

## IV. EXPERIMENTAL EVALUATIONS

We use three popular face databases to evaluate our algorithm: FERET, ORL and Face Recognition Grand Challenge version 1 experiment 1.0.4 ('FRGC-104').

### A. FERET database

In FERET database, faces are cropped to 80*64 pixels. The parameters in section II.A is: $M = 10$, $N = 4$, $S = 2$, $R = 200$. So the total dimension of the subspace is $R*M*N = 8000$, much smaller than the LGBP in [7], which is 70400.

We denote the GSF in (6) (with magnitude information) as GSF1, the GSF in (7) (without magnitude information) as GSF2. W means weighted, and IP means with illumination preprocessing. The recognition rates of different methods proposed in this paper are listed in Table I.

We can infer from Table I that: (a) When no illumination preprocessing is applied, GSF2 performs better, especially in the duplicate I and II subsets. This is probably because the Gabor magnitude is not very robust to illumination changes respectively. (b) Illumination preprocessing can improve the performance, especially in the fafc and the duplicate subsets. (c) Weighting increases the recognition rates slightly.

TABLE I. COMPARISON OF THE PROPOSED METHODS ON FERET (%)

| Method | Standard FERET probe sets | | | |
|---|---|---|---|---|
| | *fafb* | *fafc* | *dup.I* | *dup.II* |
| GSF1 | 99.2 | **95.9** | 80.0 | 56.4 |
| GSF2 | 99.3 | 93.8 | **84.9** | 69.2 |
| GSF2+W | **99.3** | 94.3 | 84.6 | **71.4** |
| IP+GSF1 | 99.4 | 99.0 | **94.3** | 89.7 |
| IP+GSF2 | 99.4 | 97.9 | 93.9 | 90.2 |
| IP+GSF1+W | **99.6** | **99.5** | 94.0 | **91.5** |

One concern about the "Binarization and joint histogram" strategy is that will the binarization step make too much discriminative information lost? So we design a new feature to see if it is better to use real value directly. First, we calculate the GMP and its derivatives, $G$, $Gx$, $Gy$ and $G_2$. $G_2 = Gxx+Gyy$. Then we evenly down-sample them to make the final feature dimension equal to the original GSF. Finally, weighted EPFDA is applied to this feature. The performance of this feature is shown in Table II. The first 3 rows correspond to the new features. As we can see, GSF shows better performance. The reason is probably that, although binarization discards some information, it also gains the robustness against illumination change. Besides, the new feature can only describes the shape of a GMP on some sample points; however, the histogram strategy in GSF summarizes the statistical property of a whole sub-region.

Table III is a comparison of our methods with other state-of-art methods. If no illumination preprocessing is done, our weighted GSF2 method outperforms LGBPHS and LGBP+EPFDA methods except the fafc subset. It is possibly because of the generalization problem of FDA. Our weighted GSF1 method with IP gets the highest scores in fafb, fafc and duplicate I, and the second highest score in duplicate II subset. It should be noticed that the last two rows of this table are the latest results on FERET, they did not use IP. However, they explored the Gabor phase information in some way. Therefore, their results are impressive. If only Gabor magnitude information is used, our result is more competitive.

### B. ORL database

ORL database is a relatively small face database; it contains 400 images corresponding to 40 subjects. We use 3 images of each subject as training and target images, others as query images. Faces in ORL have pose variance, so we need to enlarge the size of the partitioned regions. The parameters in section II are: $M = 5$, $N = 4$, $S = 1$, $R = 39$. No illumination preprocessing is used. The recognition rates are listed in Table IV.

### C. FRGC-104

The gallery images of FRGC-104 were obtained under carefully controlled conditions, meanwhile the query images were captured in uncontrolled indoor and outdoor settings. This makes the recognition task more challenging. The recognition rates are shown in Table V. Our method still outperforms the method in [8], with or without IP.

TABLE II. RECOGNITION RATES WITH OR WITHOUT THE BINARIZATION AND JOINT HISTOGRAM STRATEGY (%)

| Method | Standard FERET probe sets | | | |
|---|---|---|---|---|
| | fafb | fafc | dup.I | dup.II |
| IP+G+W (without binaization) | 99.3 | **99.5** | 89.9 | 85.9 |
| IP+G,Gx,Gy+W (without binarization) | 99.4 | 99.0 | 92.5 | 91.0 |
| IP+G,Gx,Gy,$G_2$+W (without binarization) | 99.4 | 99.0 | 92.8 | **91.5** |
| IP+GSF1+W (with binarization) | **99.6** | 99.5 | **94.0** | 91.5 |

TABLE III. COMPARISON WITH OTHER METHODS ON FERET (%)

| Method | Standard FERET probe sets | | | |
|---|---|---|---|---|
| | fafb | fafc | dup.I | dup.II |
| GSF2+W | 99.3 | 94.3 | 84.6 | 71.4 |
| LGBPHS [7] | 98.0 | 97.0 | 74.0 | 71.0 |
| LGBP+EPFDA [8] | 99.2 | 95.9 | 83.1 | 67.1 |
| IP+GSF1+W | **99.6** | **99.5** | **94.0** | 91.5 |
| IP+ LGBP+EPFDA | 99.4 | 98.5 | 93.4 | 88.5 |
| IP+GFC | 98.6 | 97.9 | 85.0 | 81.6 |
| LBP [4] | 97.0 | 79.0 | 66.0 | 64.0 |
| HGPP [10] | 97.5 | **99.5** | 79.5 | 77.8 |
| Method in [11] | 99.0 | 99.0 | **94.0** | **93.0** |

TABLE IV. THE RECOGNITION RATES ON ORL (%)

| Method | Recognition Rate |
|---|---|
| GSF1 | **97.3** |
| LGBP+EPFDA [8] | 97.0 |
| GFC [2] | 95.7 |

TABLE V. COMPARISON WITH OTHER METHODS ON FRGC-104 (%)

| Method | Recognition Rate |
|---|---|
| GSF2 | 94.9 |
| LGBP+EPFDA [8] | 94.0 |
| IP+GSF1 | **97.2** |
| IP+ LGBP+EPFDA | 96.4 |
| IP+LTP [9] | 86.3 |
| IP+GFC | 76.8 |

## V. CONCLUSION

This paper proposes a novel face representation, Gabor surface feature, which treats Gabor magnitude pictures as smooth surfaces, and extracts magnitude, $1^{st}$ and $2^{nd}$ derivative information from it. This method has a clear meaning and is easy to implement. Compared with other Gabor or LBP-based methods, experiments show that GSF surpasses them if only Gabor magnitude information is used. With illumination preprocessing and weighting strategy, it gives impressive results on FERET, ORL and FRGC-1.0.4 databases. This method actually regards Gabor filtered results as intermediate features. From this point of view, new algorithm may be designed, if one can find a way that can deeply explore the properties of Gabor filtered results.

Encouraged by the results in [10] and [11], our future work is to add Gabor phase information to GSF by analyzing the shape of Gabor phase pictures.

REFERENCES

[1] L. Shen, L. Bai, "A review on Gabor wavelets for face recognition," Pattern. Anal. & Applications, vol. 9, issue 2, pp 273-292, Sep. 2006.

[2] C. Liu and H. Wechsler, "Gabor feature based classification using the enhanced Fisher linear discriminant model for face recognition," IEEE Trans. Image Process., vol. 11, No. 4, pp. 467–476, April. 2002.

[3] T. Ojala, M. Pietikainen, and T. Maenpaa, "Multiresolution gray-scale and rotation invariant texture classification with local binary patterns," IEEE Trans. Pattern Anal. Mach. Intell., vol. 24, no. 7, pp. 971–987, July 2002.

[4] T. Ahonen, A. Hadid, and M. Pietikainen, "Face recognition with local binary pattern," in Proc. 8th Eur. Conf. Computer Vision, 2004, pp. 469–481.

[5] X. Huang, S. Z. Li, and Y. Wang, "Shape localization based on statistical method using extended local binary pattern," in Proc. International Conference on Image and Graphics, 2004, pp.184-187.

[6] Z. H. Guo, L. Zhang, and D. Zhang, "A completed modeling of local binary pattern operator for texture classification," IEEE Trans. Image Process., vol. 19, no. 6, pp. 1657–1663, June 2010.

[7] W. Zhang, S. Shan, W. Gao, X. Chen, and H. Zhang, "Local Gabor binary pattern histogram sequence (LGBPHS): A novel non-statistical model for face representation and recognition," in Proc. 10th IEEE Int. Conf. Computer Vision, 2005, pp. 786–791.

[8] S. Shan, W. Zhang, Y. Su, X. Chen, and W. Gao, "Ensemble of piecewise FDA based on spatial histograms of local (Gabor) binary patterns for face recognition," in Proc. of the 18th Int. Conf. on Pattern Recognition, vol. 4, pp. 606–609, Hong Kong, August 2006.

[9] X. Tan, and B. Triggs, "Enhanced local texture feature sets for face recognition under difficult lighting conditions," in Proc. Int. Workshop on Analysis and Modeling of Faces and Gestures, 2007, pp. 168-182.

[10] B. Zhang, S. Shan, X. Chen, and W. Gao, "Histogram of Gabor phase patterns (HGPP): a novel object representation approach for face recognition," IEEE Trans. on Image Processing, vol. 16, No. 1, pp. 57-68, January 2007.

[11] S. Xie, S. Shan, X. Chen, and J. Chen, "Fusing local patterns of Gabor magnitude and phase for face recognition," IEEE Trans. on Image Processing, vol. 19, No. 5, pp. 1349-1361, May 2010.